\documentclass[conference]{IEEEtran}
\IEEEoverridecommandlockouts

\usepackage{cite}
\usepackage{amsmath,amssymb,amsfonts}
\usepackage{algorithmic}
\usepackage{graphicx}
\usepackage{textcomp}
\usepackage{xcolor}

\usepackage{graphics} 
\usepackage{epsfig} 
\usepackage{mathptmx} 
\usepackage{times} 
\usepackage{siunitx}
\usepackage{hyperref}

\UseRawInputEncoding
\usepackage[capitalise]{cleveref}
\usepackage{subcaption}
\usepackage{acro}

\usepackage{algorithm}
\usepackage{mathtools}
\usepackage{fancyhdr}

\def\BibTeX{{\rm B\kern-.05em{\sc i\kern-.025em b}\kern-.08em
    T\kern-.1667em\lower.7ex\hbox{E}\kern-.125emX}}

\fancypagestyle{FirstPage}{
	\lfoot{{\small \copyright2026 IEEE. Personal use of this material is permitted. Permission from IEEE must be obtained for all other uses, in any current or future media, including reprinting/republishing this material for advertising or promotional purposes, creating new collective works, for resale or redistribution to servers or lists, or reuse of any copyrighted component of this work in other works.}}
}

\DeclareAcronym{bev}{
	short=BEV, 
	long=Bird's Eye View
}

\DeclareAcronym{GT}{
	short=GT, 
	long=Ground Truth
}

\DeclareAcronym{LR}{
	short=LR, 
	long=Learning Rate
}

\DeclareAcronym{TP}{
	short=TP, 
	long=True Positive
}

\DeclareAcronym{hd}{
	short=HD, 
	long=high-definition
}
\DeclareAcronym{iou}{
	short=IoU, 
	long=Intersection over Union
}
\DeclareAcronym{miou}{
	short=mIoU, 
	long=mean Intersection over Union
}
\DeclareAcronym{semseg}{
	short=sem-seg, 
	long=semantically segmented
}

\begin{document}

\title{Faster Training, Fewer Labels: Self-Supervised Pretraining for Fine-Grained BEV Segmentation\\












	\thanks{$^{1}$University of Wuppertal, $^{2}$APTIV, daniel.busch2@uni-wuppertal.de}
}
\author{\IEEEauthorblockN{Daniel Busch$^{1, 2}$ }
	\and
	\IEEEauthorblockN{Christian Bohn$^{1, 2}$ }
	\
	\and
	\IEEEauthorblockN{Thomas Kurbiel$^2$ }
	\and
	\IEEEauthorblockN{Klaus Friedrichs$^2$ }
	\and
	\IEEEauthorblockN{Richard Meyes$^1$ }
	\and
	\IEEEauthorblockN{Tobias Meisen$^1$ }
}


\maketitle
\thispagestyle{FirstPage}
\begin{abstract}
    Dense \ac{bev} semantic maps are central to autonomous driving, yet current multi-camera methods depend on costly, inconsistently annotated BEV ground truth.
    We address this limitation with a two-phase training strategy for fine-grained road marking segmentation that removes full supervision during pretraining and halves the amount of training data during fine-tuning while still outperforming the comparable supervised baseline model.
    During the self-supervised pretraining, BEVFormer predictions are differentiably reprojected into the image plane and trained against multi-view semantic pseudo-labels generated by the widely used semantic segmentation model Mask2Former. A temporal loss encourages consistency across frames.
    The subsequent supervised fine-tuning phase requires only $50\%$ of the dataset and significantly less training time. With our method, the fine-tuning benefits from rich priors learned during pretraining boosting the performance and BEV segmentation quality (up to $+2.5 pp$ \ac{miou} over the fully supervised baseline) on nuScenes. It simultaneously halves the usage of annotation data and reduces total training time by up to two thirds. The results demonstrate that differentiable reprojection plus camera perspective pseudo labels yields transferable BEV features and a scalable path toward reduced-label autonomous perception.
\end{abstract}

\section{Introduction}
\label{sec:intro}

\begin{figure*}[h]
      \centering
      \includegraphics[width=0.98\textwidth]{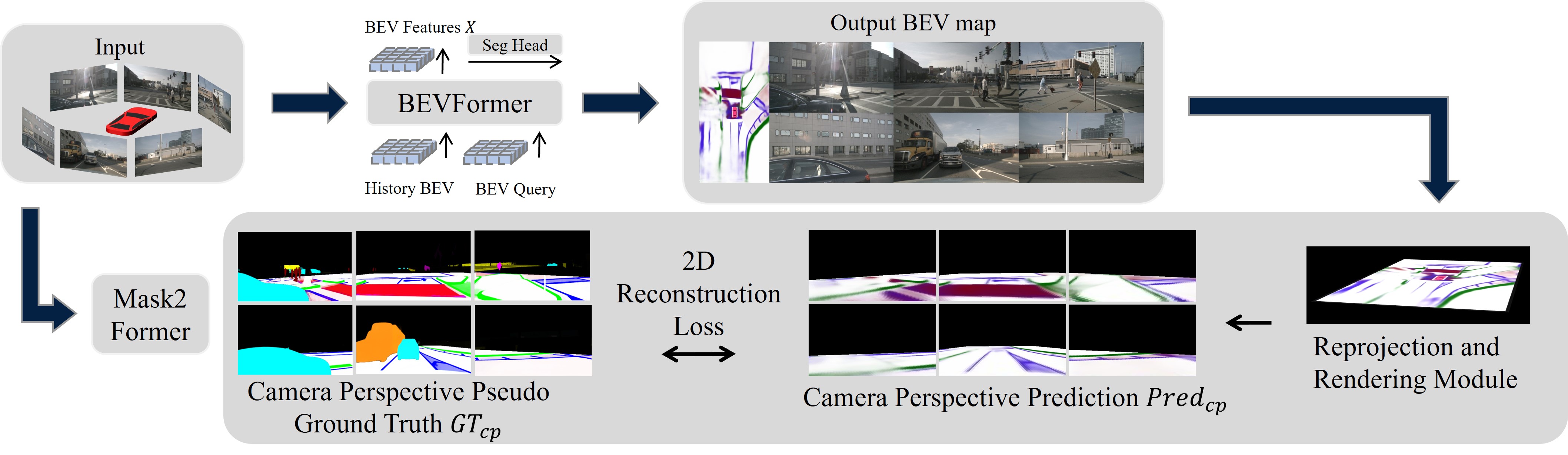}
      \caption{BEVFormer architecture \cite{li_bevformer_2022} with the self-supervised pretraining. Left: Camera input. Center: the BEVFormer with its transformer module containing Temporal Self-Attention into History \ac{bev} and spatial Cross-Attention. Right: Output of the BEVFormer. Bottom right the differentiable reprojection and rendering. Bottom center and left 2D Reconstruction loss comparing the six reconstructed predctions $Pred_{cp}$ with the camera perspective pseudo ground truth $GT_{cp}$.}
      \label{architecture}
\end{figure*}

\acf{bev} maps are a crucial representation for perception in ADAS and autonomous driving systems. They enable a unified view of road layouts, dynamic objects, and scene semantics, which is essential for downstream tasks such as planning and control.
Current approaches to BEV generation as presented in \cite{ozturk_glane3d_2025,li_dualbev_2025,le_diffusion_2025,choi_mask2map_2025,chang_recurrentbev_2025,liu_bevfusion_2024,li_bev-lgkd_2024,chambon_pointbev_2024,leng_bevcon_2025} rely heavily on supervised learning with manually labeled BEV ground truth.
Such labels are costly to produce, difficult to maintain across large areas \cite{bao_high-definition_2022}, and often inconsistent between datasets \cite{ozturk_glane3d_2025,caesar_nuscenes_2020,fong_panoptic_2022,sun_scalability_2020,liao_kitti-360_2022}. This limits the scalability of BEV-based methods and hampers their ability to generalize to new environments.

What remains underexplored, especially for fine-grained structures such as road markings, is reducing the reliance on dense BEV supervision. We present a training strategy from multi-view camera images (without LiDAR or depth ground truth), that surpasses fully supervised methods yet requires only half of the BEV ground truth and half of training steps.


We propose a two phase training strategy that comprises a self-supervised pretraining with a reduced supervised fine-tuning. The underlying baseline model is the BEVFormer \cite{li_bevformer_2022} that learns to generate BEV maps directly from multi-view camera images using an encoder-decoder architecture.
As depicted in \cref{architecture}, instead of relying on BEV ground truth for the self-supervised pretraining, our method reprojects the predicted BEV segmentation map back into the image plane and supervises it with semantic pseudo-labels in camera perspective. A differentiable rendering module enables end-to-end optimization, while a temporal consistency loss encourages stable predictions across consecutive frames. The subsequent, supervised fine-tuning benefits from the pretraining and can focus on aligning with the nuScenes ground truth.
Our main contributions to support our training paradigm are:
\begin{itemize}
      \item A novel self-supervised pretraining framework for BEV segmentation that removes the requirement for BEV ground truth;
      \item A differentiable rendering pipeline that reprojects BEV predictions into the image space for supervision;
      \item A temporal loss that enforces consistency across frames, improving robustness;
      \item A two phase training strategy consisting of self-supervised pretraining and of-the-shelf supervised fine-tuning enabling the comparison with fully supervised trainings;
      \item Extensive experiments showing that our two phase method outperforms the fully supervised baseline while requiring less training time and less labels.
\end{itemize}

Our approach halves the dependence on costly BEV annotations while improving performance by +2.2pp mIoU, and additionally providing options with superior results that require a third of the total training steps making BEV segmentation more scalable and adaptable.

\section{Related Works}
\label{sec:related_works}
\subsection{Supervision strategies for Multi-Camera \ac{bev} prediction}


Building upon early works such as Lift, Splat, Shoot \cite{philion_lift_2020}, modern state-of-the-art surround-view methods ingest multi-camera surround imagery to produce \ac{bev} predictions. These approaches broadly fall into two categories: Firstly 3D object (and lane) detection, and secondly dense semantic \ac{bev} segmentation.
Differing from our approach, the first category of models as presented in \cite{ozturk_glane3d_2025,li_dualbev_2025,chang_recurrentbev_2025,leng_bevcon_2025,li_fb-bev_2023} focus on detection and localization of object in the \ac{bev} space. These models require datasets that provide 3D bounding boxes, or in case of lane detection as depicted in \cite{choi_mask2map_2025,chambon_pointbev_2024} poly-lines as ground truth.
Our approach falls within the second category of models, representative works \cite{teng_360bev_2024, le_diffusion_2025,liu_bevfusion_2024, yang_bevformer_2023,xie_m2bev_2022,harley_simple-bev_2023} predict dense semantically segmented \ac{bev} maps. These models depend on datasets with dense semantic annotations in the \ac{bev} space as ground truth. Our method however, requires this only for the fine-tuning phase.\\
In addition to that, some models require both 3D bounding boxes and dense semantic \ac{bev} maps for ground truth \cite{le_diffusion_2025,xie_m2bev_2022,li_bevformer_2022,harley_simple-bev_2023} as they predict 3D objects and semantically segmented \ac{bev}.
Contrary to our framework, some models rely on LiDAR point clouds as supervision signal. In \cite{li_bev-lgkd_2024} these point clouds are used to guide a teacher-student training process. Furthermore, in \cite{gosala_birds-eye-view_2022} panoptic segmentation labels as ground truth \ac{bev} maps are generated from point clouds.\\
Moving further to already existing self-supervised approaches, in \cite{gosala_skyeye_2023} a pseudo-labeling pipeline to generate ground truth from unlabeled data is introduced similar to our pretraining approach. Despite this, their approach generates pseudo ground truth \ac{bev} maps in a multistep process. As this process is responsible for the training objective the approach is strongly dependent on the accuracy of each step including depth estimation, semantic segmentation, and forward projection. In the successor \cite{leonardis_letsmap_2025} a self-supervised pretraining strategy based on RGB reconstruction and depth map prediction is used whereas our approach intends to predict \ac{bev} maps already in the pretraining resulting in rich \ac{bev} representations which reduces the distribution shift between pretraining and fine-tuning.
Similar to our pretraining phase, in \cite{monteagudo_rendbev_2025} semantically segmented images in the camera perspective are utilized. In contrast, their approach employs a different model \cite{hayler_s4c_2023} to generate camera-perspective pseudo ground-truth for a subsequent frame, enforcing time consistency in the forward path. Furthermore, they are only using the single front facing camera. Moreover, the three aforementioned models \cite{gosala_skyeye_2023,leonardis_letsmap_2025,monteagudo_rendbev_2025} did not show their performance on fine-grained semantic classes like lane dividers as they are trained to predict more coarse classes like drivable area and sidewalks \cite{liao_kitti-360_2022}. These limitations are addressed in our approach by utilizing a surround-view camera setup which comes with additional challenges and opportunities like stitching the semantic information from multiple views into a coherent \ac{bev} map and sensing the full $360^\circ$ environment. Furthermore, we focus on road markings by utilizing the Mask2Former model \cite{cheng_masked-attention_2022} to generate pseudo-labels in the camera perspective.

\section{Method}
The proposed method is illustrated in \cref{architecture}.
Our pretraining method utilizes the whole training pipeline of the BEVFormer \cite{li_bevformer_2022} to predict semantically segmented \ac{bev} maps. Subsequently, we extend that with the reprojection, rendering and 2D reconstruction loss (bottom half of \cref{architecture}) to enable self-supervised pretraining. We reproject the predicted \ac{bev} map $Pred_{bev}$ back into the image plane and supervise it with 2D semantic labels, the camera perspective pseudo Ground Truth $GT_{cp}$. There are multiple reasons for this design choice: First, this allows us to leverage readily available pretrained 2D segmentation models to generate the pseudo ground truth. Second, this pseudo GT generation in camera perspective does not require complex perspective transformation. Lastly, the reprojection from \ac{bev} space into the camera perspective is a 3D to 2D operation and thus mathematically well-defined. Additionally, we introduce a temporal consistency loss as part of the 2D reconstruction loss. This loss enforces stable predictions across frames to effectively utilize temporal information and mitigate occlusion issues in the camera perspective supervision. The following sections describe the BEVFormer (\cref{basemodel}), the extended architecture (\cref{pre_architecture}), supervision strategy (\cref{extension}), and temporal consistency objective (\cref{temporalloss}) in detail.

\subsection{Base Model}\label{basemodel}
The foundation of this work is the BEVFormer as introduced in \cite{li_bevformer_2022,bin-ze_bevformer_segmentation_detection_2023}, a model for full $360^\circ$ surround view \ac{bev} segmentation and 3D object detection. It is designed to process multiple camera images by incorporating two attention mechanisms based on the transformer module of the Deformable Detection Transformer (DETR) \cite{zhu_deformable_2021}. A spatial cross attention that attends to the different camera perspectives and a temporal attention that recurrently integrates past \ac{bev} features with the current ones using a temporal attention mechanism. In more detail, the \ac{bev} features from past samples are stored and motion compensated forming a history \ac{bev} in the latent space. The temporal attention can attend into this history \ac{bev} following the mechanisms adapted from DETR. The \ac{bev} features $X$, are passed to task-specific output heads to perform 3D object detection and semantic segmentation.
The only adaption made to the original BEVFormer is the focus on semantic segmentation as the sole task since the self-supervised detection of 3D objects introduces other challenges we will address in future work.

\subsection{Pretraining Architecture}\label{pre_architecture}
Rather than modifying the BEVFormer architecture itself, we append our extensions starting with the reprojection and rendering module to the training pipeline as illustrated in \cref{architecture} (Rendering Module). In this way, the BEVFormer and our extensions are kept modular with the predicted \ac{bev} map as the interface between both parts. This enables the simple transition to the fine-tuning phase without the extensions and makes the BEVFormer interchangeable for future work as the parts didn't merge.\\
BEVFormer natively supports surround-view camera input and incorporates temporal information from past frames. This is particularly beneficial for \ac{bev} semantic segmentation as it allows handling occlusion by dynamic objects and improves the overall segmentation quality \cite{li_bevformer_2022}.\\
The appended reprojection and rendering module takes the predicted \ac{bev} map $Pred_{bev}$, projects it onto a ground plane mesh in the 3D vehicle coordinate system and renders it into the camera perspective using a differentiable rendering module \cref{implementation_details}.
Using the predicted \ac{bev} map $Pred_{bev}$ rather than intermediate features $X$, places it directly on the computation graph. It is therefore optimized by backpropagation without an additional auxiliary loss.\\
In more detail, the ground plane mesh is identical to the extend of the query grid of the BEVFormer and defined by its four corner coordinates. The mesh consists of two triangles and is textured with $Pred_{bev}$ which is treated as a multi color-channeled image.
This scene construction is motivated by the fact that road markings are located on the ground plane and the texture can take the semantic information as the multichannel image $Pred_{bev} \in \mathbb{R} ^{l\times w_{bev} \times c }$, where $l$ and $w_{bev}$ denote the spatial dimension, and $c$ represents the number of semantic classes. This format is directly provided by the BEVFormer output.
The subsequent rendering procedure results in six rendered images, forming the \textbf{c}amera \textbf{p}erspective predictions $Pred_{cp} \in \mathbb{R} ^{6\times h_{img} \times w_{img} \times c}$, one for each of the six camera perspectives, with spatial dimension $h_{img} \times w_{img}$. Each prediction is provided as one-hot encoded vector, which is a common format for loss computation in semantic segmentation tasks.

\begin{figure}[htbp]
    \centering
    \begin{subfigure}{0.95\linewidth}
        \parbox[b][1.5cm][l]{0.05\textwidth}{
            \caption{}
            \label{lanethicknessa}
        }
        \parbox[b]{\textwidth}{
            \rotatebox[origin=l]{32}{}\includegraphics[width=0.42\linewidth]{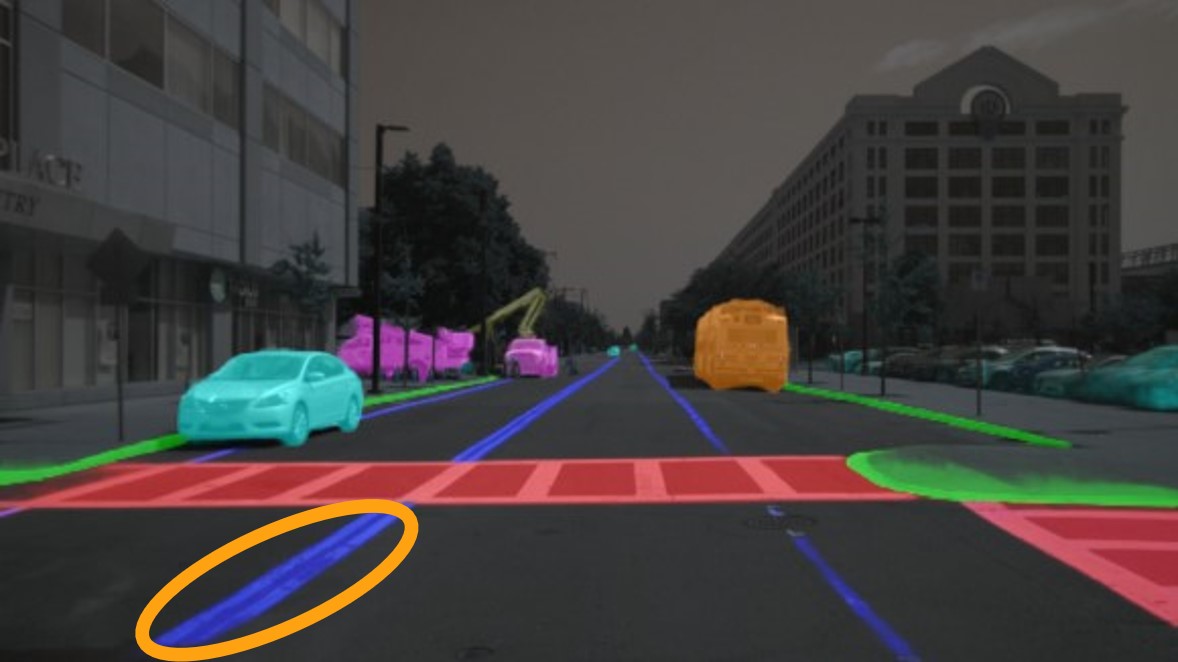}
            \rotatebox[origin=l]{62}{}\includegraphics[width=0.52\linewidth]{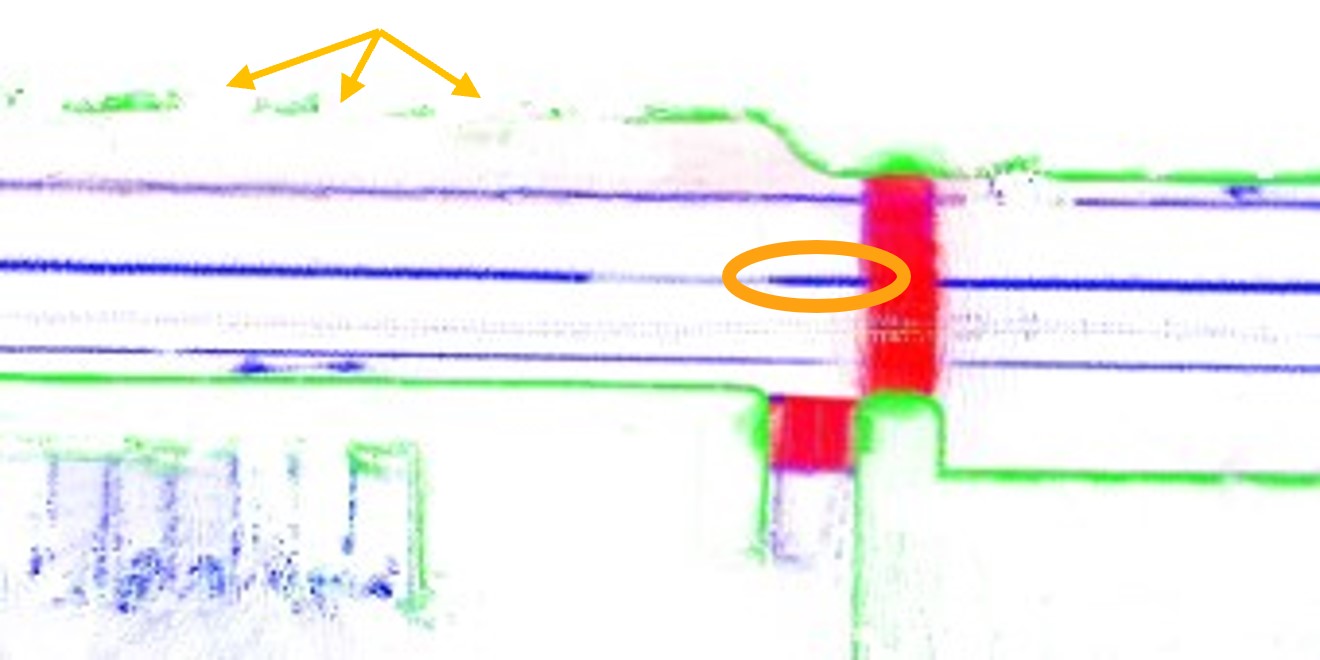}
        }

    \end{subfigure}

    \begin{subfigure}{0.95\linewidth}
        \parbox[b][1.5cm][l]{0.05\textwidth}{
            \caption{}
            \label{lanethicknessb}
        }
        \parbox[b]{\textwidth}{
            \rotatebox[origin=l]{32}{}\includegraphics[width=0.42\linewidth]{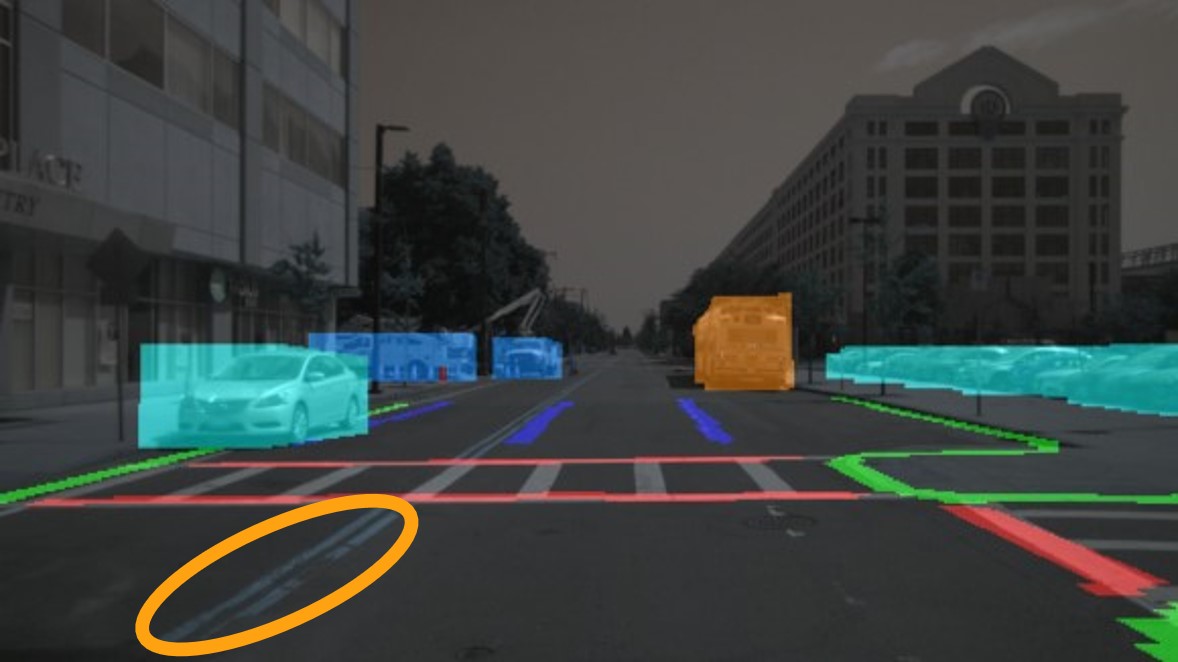}
            \rotatebox[origin=l]{62}{}\includegraphics[width=0.52\linewidth]{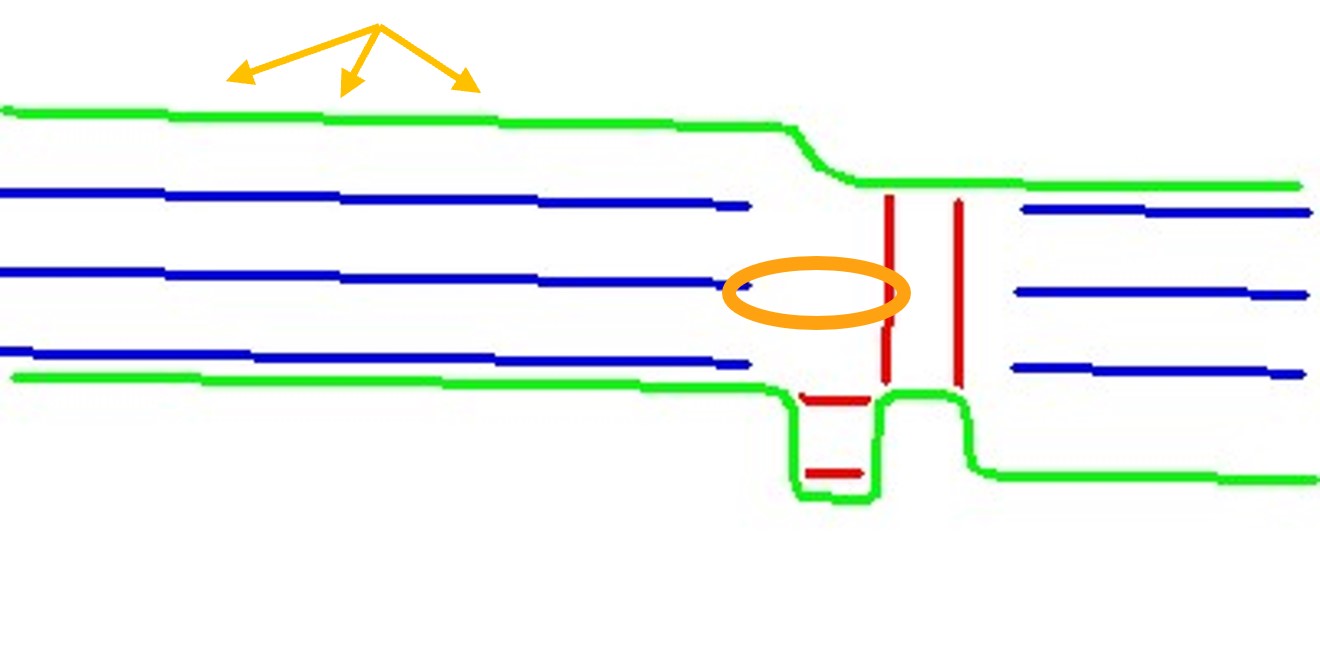}

        }
    \end{subfigure}
    \caption{Ground truth comparison for pseudo ground truth (a) and NuScenes ground truth (b) in front view (left) and \ac{bev} (right). For comparison the pseudo ground truth is projected to \ac{bev} using nuScenes lidar data. (\textcolor{green}{road boundary}; \textcolor{blue}{lanes}; \textcolor{red}{crosswalk}).}
    \label{lanethickness}

\end{figure}

\subsection{Supervision Strategies}\label{extension}
In this section, we detail the generation of the camera perspective pseudo ground truth $GT_{cp}$ for the self-supervised pretraining phase and outline the fine-tuning phase.
\\
\textbf{Pseudo GT.} To supervise the camera perspective predictions $Pred_{cp}$, we need to automatically transform the camera input, into a camera perspective pseudo ground truth.
For this purpose, we leverage the Mask2Former model \cite{cheng_masked-attention_2022} which is able to perform high-quality semantic segmentation on single images and is trained on the Mappilary Vistas dataset \cite{neuhold_mapillary_2017}. This dataset contains a wide variety of driving scenes with high-quality pixel-wise annotations following similar classes types than nuScenes, making it suitable for our task. Through this process, we obtain semantically segmented images (\cref{architecture}, Pseudo Ground Truth) including three road marking classes (lane dividers, road boundaries, and crosswalks) and ten object classes that frequently occlude road markings in the camera perspective. To give an intuition of the segmentation quality, we provide a comparison on one example output for both the pseudo ground truth (\cref{lanethicknessa}) and the nuScenes ground truth (\cref{lanethicknessb}). Despite the fact that both ground truth types are not perfect, containing errors and differences in labeling styles, both contain the same class types. Thus, in terms of channel dimension and class types the pretrained model can build a suitable camera-perspective pseudo ground truth $GT_{cp} \in \mathbb{R} ^{6\times h_{img} \times w_{img} \times c }$. We evaluate the influence of the differences between these annotations by comparing our method with a supervised baseline in \cref{experimants}.\\
This method of annotation generation can eliminate the reliance on direct \ac{bev} ground truth and presents a notable advantage, as this form of pseudo ground truth can be generated automatically using pretrained models like \cite{cheng_masked-attention_2022, kirillov_segment_2023} and is thus more scalable to produce large quantities of training data.\\
\textbf{Fine-tuning} The fine-tuning phase uses the standard nuScenes \ac{bev} ground truth maps for supervision, following the original BEVFormer training procedure \cite{li_bevformer_2022}. We utilize this straightforward fine-tuning without additional architectural modifications as our focus lies on the pretraining phase, that aims to already provide a useful \ac{bev} representation. Furthermore, this phases is reduced in terms of \ac{bev} ground truth labels and training steps as detailed in \cref{experimants}.\\

\subsection{2D Reconstruction and Temporal Loss}\label{temporalloss}



The rendered camera perspective predictions $Pred_{cp}$ are supervised using the camera perspective pseudo ground truth $GT_{cp}$. As both, predictions and ground truth, are in the same format, we can directly apply a pixel-wise loss function. We define a 2D reconstruction based on a cross-entropy loss which is widely used in semantic segmentation tasks and follows the same principle as in the original BEVFormer \cite{li_bevformer_2022}. The loss function is formulated as a pixel-based cross-entropy $L_{CE}$ loss over all classes:
\begin{equation}\label{crosentropy}
    L_{CE} =\frac{-\sum_{\text{pixels}}GT_{cp} \cdot \text{LogSoftmax}(Pred_{cp})}{n_{\text{pixels}}}
\end{equation}
where $n_{\text{pixels}}$ denotes the total number of pixels across all channels and views.\\

\subsubsection{Temporal Loss}\label{temporalloss_sub}
In the original supervised setting, the ground truth maps provide the complete road layout of the surrounding scene without any occlusions. This allows the model to learn the complete road structure, even if parts are occluded by dynamic objects. Missing information may be found in the history \ac{bev} using the temporal attention as described in \cref{basemodel}. In contrast, the camera perspective pseudo ground truth only contains road markings currently visible to the ego-vehicle. Thus, if a road marking is occluded in the current frame, the model has no incentive to retain this information in the latent \ac{bev} features $X$ as it is not part of the current supervision. This may lower the quality of the predicted \ac{bev} maps.

\begin{figure}[htbp]
    \centering
    \begin{subfigure}{0.64\linewidth}
        \centering
        \includegraphics[width=.8\linewidth]{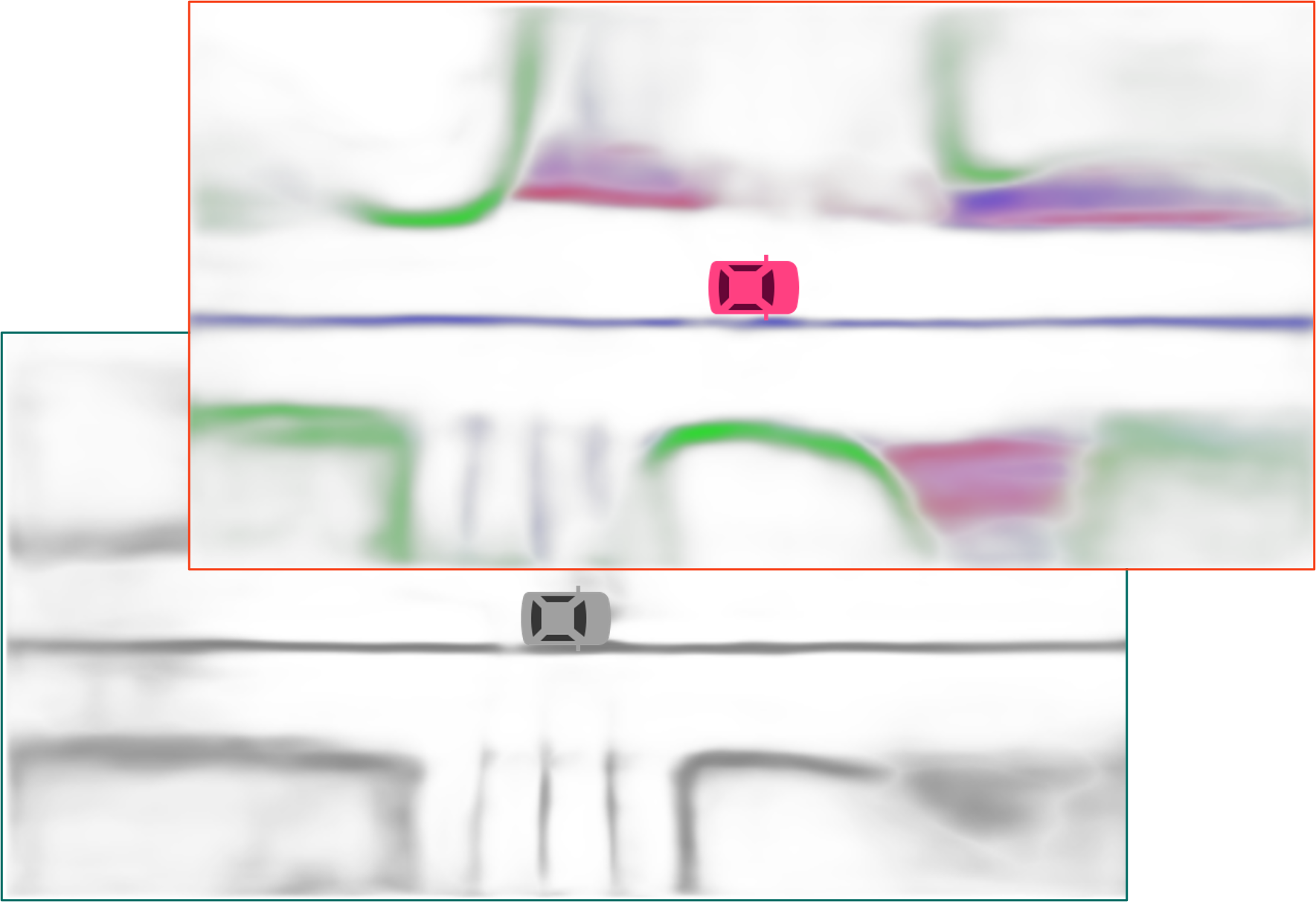}
        \caption{Predicted \ac{bev} sample of ego motion compensation from $t$ (colorized) to the past time step $t-1$ (grayscale)}
    \end{subfigure}\label{temploss}
    \begin{subfigure}{0.32\linewidth}
        \includegraphics[width=\linewidth]{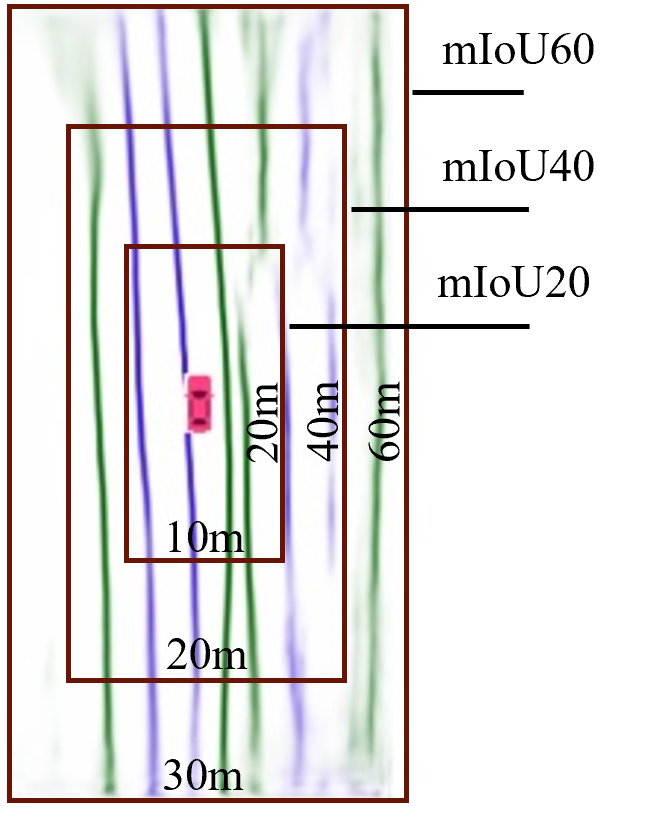}
        \caption{Visualization of the different \acp{miou} with their corresponding area}
    \end{subfigure}\label{miou}
\end{figure}

\begin{algorithm}[htbp]
    \caption{Temporal loss algorithm}
    \label{motioncompalgo}
    \begin{algorithmic}
        \REQUIRE{Ego movement between $t$ and $t-1$: $\Delta x$, $\Delta y$ $\Delta \varphi$, latent \ac{bev} features $X$}
        \STATE $ X_{rot} \leftarrow \text{rotate}(X, \Delta \varphi)$
        \STATE $ X(t-1) \leftarrow \text{translate}( X_{rot}, \Delta x, \Delta y)$
        \FOR{i $\in$ $\{0, 1\}$}
        \STATE $Pred_{bev}(t-i) = $SegHead$(X(t-i))$
        \STATE $Pred_{cp}(t-i) = $ Diff. RenderModule$(Pred_{bev}(t-i))$\\
        \STATE $L_{CE}(t-i) = $ CELoss$(Pred_{cp}(t-i),GT_{cp}(t-i))$\\
        \ENDFOR
        \RETURN $L_{CE}, L_{CE}(t-1)$
    \end{algorithmic}
\end{algorithm}
To force the model to effectively utilize the temporal context we predict the \ac{bev} map and compute the 2D reconstruction loss for both the current and the previous time step based on the current latent \ac{bev} features $X$. Considering the ego movement between two time steps $t$ and $t-1$ we implement an ego motion compensation as depicted in \cref{temploss}. This compensation is to a large extent the inverse operation of the compensation in the temporal attention mechanism described in \cite{li_bevformer_2022}. The algorithm for both losses is summarized in \cref{motioncompalgo}.\\
We choose the immediately preceding frame $t-1$ for the temporal loss as a trade-off: On the one hand the overlapping visible areas between both time steps is higher for smaller time gaps.
On the other hand, a larger time gap increases the likelihood of dynamic objects moving away from occluded road markings, revealing them in the past frame. Considering the nuScenes frame rate of 2 Hz and a usual urban ego motion of $30 - 50 km/h$ the recording positions already differ by circa $4 - 7 m$.\\
Accordingly, the model is encouraged to retain enough information from past frames in the current latent \ac{bev} features $X$ to predict also the past \ac{bev} map and camera perspective prediction $Pred_{cp}(t-1)$. To some extent, the model can learn to fill in occluded road markings from the past frame into the current \ac{bev} features $X$, due to this cross-coupling. Importantly, we use the same segmentation head for both time steps, ensuring that the additional loss term $L_{CE}(t-1)$ is directly linked to the current prediction by sharing the same weights.\\

\begin{table*}[htbp]
	\centering
	\vspace{0.2cm}
	\resizebox{0.85\textwidth}{!}{
		\parbox[c][2cm][c]{0.99\textwidth}{}
		\begin{tabular}{l | l l l |r r r}
			\hline
			Method                       & \begin{tabular}{@{}c@{}}boundary\\IoU $\uparrow$\end{tabular} & \begin{tabular}{@{}c@{}}lane\\IoU $\uparrow$\end{tabular} & \begin{tabular}{@{}c@{}}crosswalk \\IoU $\uparrow$ \end{tabular} & \begin{tabular}{@{}c@{}}mIoU20 $\uparrow$ \end{tabular} & \begin{tabular}{@{}c@{}}mIoU40 $\uparrow$ \end{tabular} & \begin{tabular}{@{}c@{}}mIoU60 $\uparrow$ \end{tabular} \\[0.5ex]
			\hline
			supervised baseline          & 25.1                      & 26.4                      & 11.5                      & 31.5                      & 25.6                      & 21.0                      \\
			self-supervised pretraining  & 7.1                       & 6.0                       & 3.2                       & 6.3                       & 6.0                       & 5.4                       \\
			self-superv. pretrain + temp & 6.8                       & 6.0                       & 5.4                       & 7.4                       & 7.0                       & 6.1                       \\
			\hline

			ours(3ep pretrain + temp)    & 25.2                      & 27.2                      & 14.9                      & 32.7                      & 26.7                      & 22.4                      \\
			ours(3ep pretrain)           & 24.8                      & 27.0                      & 14.9                      & 32.7                      & 26.7                      & 22.3                      \\
			ours(5ep pretrain + temp)    & 24.5                      & 26.4                      & 14.8                      & 32.3                      & 26.3                      & 21.9                      \\
			ours(10ep pretrain + temp)   & 26.2                      & 27.2                      & 15.3                      & 33.5                      & 27.5                      & 22.9                      \\
			ours(20ep pretrain + temp)   & 26.2                      & \underline{27.7}          & 15.6                      & \underline{35.3}          & 28.2                      & 23.2                      \\
			ours(22ep pretrain + temp)   & \underline{26.3}          & \textbf{28.1}             & \underline{16.0}          & 34.5                      & \underline{28.6}          & \textbf{23.5}             \\
			ours(22ep pretrain)          & 26.4                      & 27.3                      & 15.3                      & 33.5                      & 27.8                      & 23.3                      \\
			ours(24ep pretrain + temp)   & 26.2                      & \textbf{28.1}             & \underline{16.0}          & 35.1                      & 28.5                      & \underline{23.4}          \\
			ours(26ep pretrain + temp)   & 25.8                      & \underline{27.7}          & \textbf{16.4}             & 34.7                      & 28.3                      & 23.3                      \\
			ours(27ep pretrain + temp)   & \textbf{26.5}             & 27.3                      & 15.7                      & \textbf{35.8}             & \textbf{28.7}             & 23.2                      \\
			\hline
		\end{tabular}}
	\caption{The different \acp{iou} for our methods: Self-superv. pretrain: supervision with camera perspective pseudo ground truth, temp: Temporal loss mechanism included, ours($x$ ep pretrain): two phase training with $x$ epochs pretraining and fine-tuning on the halved dataset. mIoU20: $20m$ consideration range ($1/3$), mIoU40: $40m$ consideration range ($2/3$), mIoU60: full range of $60m$. $\uparrow$: higher values are better. \textbf{best results}, \underline{second best results}.}
	\label{pretraining_length}
\end{table*}
\section{Data and Implementation}\label{dataset}
We use the nuScenes dataset \cite{caesar_nuscenes_2020} for our experiments as it provides the semantic map classes: road boundary, lane divider, and crosswalk. The dataset, widely adopted in recent state-of-the-art studies \cite{leonardis_letsmap_2025,li_dualbev_2025,leng_bevcon_2025,li_bevformer_2022}, comprises $1{,}000$ scenes, each $20$ seconds long, collected in Boston and Singapore. Each scene includes data from six cameras, five radars, and one LiDAR sensor. The cameras cover a full 360-degree view around the vehicle. The dataset includes annotations for 23 object classes and eight semantic map classes. For our experiments, we focus on the semantic map classes which are used in the original BEVFormer: road boundary, lane divider, and crosswalk.
We use the input images of the full dataset for our self-supervised pretraining consisting of 28,130 training samples and 6,019 validation samples. For the supervised fine-tuning we reduce the training set to $50\%$ of the original training set, resulting in $14{,}065$ training samples as this still yields to result improvement. To reduce computational costs, we reduce the image resolution from $1600 \times 900$ pixels to $512 \times 288$ pixels. This allows us to train our model with the reprojection and rendering module on a single NVIDIA A100 GPU and fulfill a requirement of the BEVFormer architecture that the image width are divisible by 32.

\subsection{Implementation details}\label{implementation_details}
We closely follow the implementation details of the original BEVFormer \cite{li_bevformer_2022}. For the backbone, we use a pretrained ResNet-50 \cite{he_deep_2016} model. The query grid is sized to $150 \times 150$ with a range of $[-15m, 15m]$ in the $X$ direction and $[-30m, 30m]$ in the $Y$ direction. This setup corresponds closely to the of-the-shelf BEVFormer in medium size as presented in \cite{bin-ze_bevformer_segmentation_detection_2023}.
The reprojection and rendering module is implemented using PyTorch3D components \cite{ravi_accelerating_2020}.
It provides components to build scenes in the 3D space, perform transformations, and render the scene from different camera perspectives. We adapt these components to our method to build the \ac{bev} in the 3D vehicle coordinate system and perform rendering by first translating and rotating the 3D mesh scene according to the extrinsic camera parameters of each of the six cameras. Lastly, rasterization is performed to obtain the camera perspective predictions $Pred_{cp}$.\\
For pseudo ground-truth generation we employ the publicly released Mask2Former model \cite{cheng_masked-attention_2022} pretrained for $300{,}000$ iterations on the Mapillary Vistas dataset \cite{neuhold_mapillary_2017}, achieving $57.4$ \ac{miou} as reported in \cite{cheng_masked-attention_2022}. We adopt the compact configuration with a ResNet-50 backbone \cite{he_deep_2016}, which delivers sufficient segmentation quality to pass our baseline while reducing computational cost relative to larger variants.\\

\subsection{Evaluation metrics}\label{evaluation_metrics}
\begin{figure}[h]
	\centering
	\includegraphics[width=0.9\linewidth]{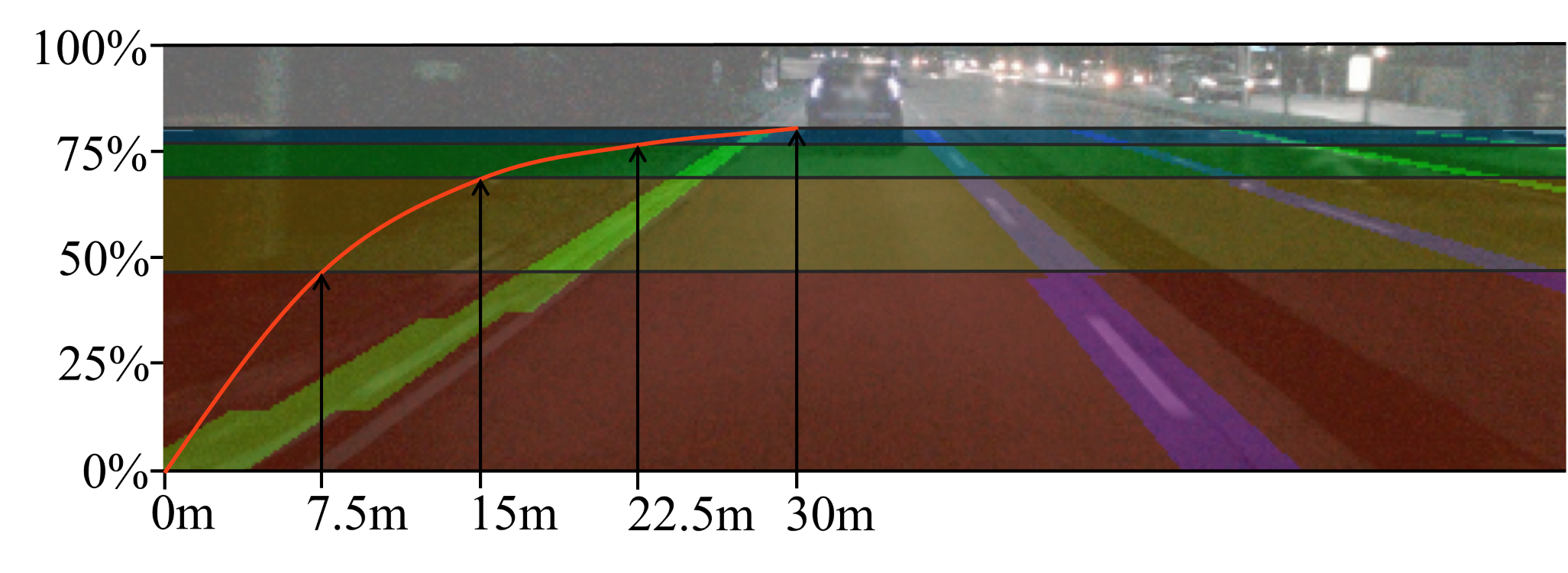}
	\caption{Relation between road distance in $m$ and image portion in $\%$. The different colors represent equally long distances each covering one quarter of the total predicted distance.}
	\label{pixelrelation}
\end{figure}

To evaluate the predicted \ac{bev} maps, we use \ac{iou} metrics for each class and the \ac{miou} metric as these metrics are commonly used to evaluate semantic segmentation tasks e.g. in \cite{li_bevformer_2022,leonardis_letsmap_2025,monteagudo_rendbev_2025}. Additionally, they are supported by the nuScenes dataset \cite{caesar_nuscenes_2020}. The \ac{miou} is calculated as the average over the \acp{iou} of all classes.
Additionally, we evaluate over different distance ranges as depicted in \cref{miou}. This subdivision is chosen based on the observation that the distances in the \ac{bev} plane are not evenly distributed in the camera perspective. The closer the distance, the more pixels are available in the image. As shown in \cref{pixelrelation}, the first $7.5m$ of the road are represented by approximately $50\%$ of the image, whereas the $7.5m$ from $22.5m$ to $30m$ by circa $6\%$. Therefore, we evaluate the \ac{miou} in three different ranges: $20m$ (1/3 of the total range), $40m$ (2/3 of the total range), and $60m$ (full range). This allows us to assess the model's performance at different distances from the vehicle.

\section{Experiments}\label{experimants}
We observe that our baseline model, the original BEVFormer, starts to converge at $30$ epochs which corresponds to $140{,}600$ steps with a batch size of six.
Our method is composed of two training phases, the self-supervised pretraining and the supervised fine-tuning.
We reduce the epoch length of the fine-tuning phase to half the original length, requiring only half of the \ac{bev} ground truth maps. This is due to the observation that the supervised fine-tuning phase still provides improvements using only $50\%$ of the training data.
As the two training phases have different epoch lengths we report the total training length in steps to maintain comparability. We set the fine-tuning phase to $32{,}800$ steps (14 halved epochs), since we focus on the influence of the pretraining length. We have observed that first signs of convergence appear for the majority of our testes within these $32{,}800$ steps.

To validate the influence of the temporal loss mechanism and the pretraining length, we conduct two ablation studies as described in the following sections.
For this purpose we pretrain with and without the temporal loss mechanism for the full training length of $140{,}600$ steps and compare it against the baseline. The results are shown in the upper part of \cref{pretraining_length}.
In the main part of \cref{pretraining_length} the second ablation study of our two phase training approach is presented with different pretraining lengths between $3$ and $27$ epochs while keeping the fine-tuning length constant at $32{,}800$ steps.
For the first ablation study the pretraining is done for $30$ epochs. Consequently, each pseudo ground truth sample is needed $30$ times during training. To optimize the training process, we generate the pseudo ground truth offline before training. In doing so, we can save computational resources and time, as the pseudo ground truth only needs to be generated once. With this setup, we are able to train the model with the reprojection and rendering module on a single NVIDIA A100 GPU.

\begin{figure}[htbp]
	\centering
	\begin{subfigure}[t]{0.24\textwidth}
		\centering
		\rotatebox[origin=b]{270}{
			\includegraphics[width=0.5\linewidth]{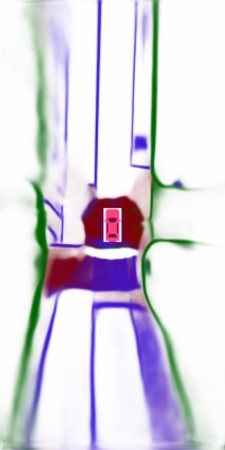}}
		\caption{pretrain} \label{img:16_ssl}
	\end{subfigure}
	\begin{subfigure}[t]{0.24\textwidth}
		\centering
		\rotatebox[origin=b]{270}{
			\includegraphics[width=0.5\linewidth]{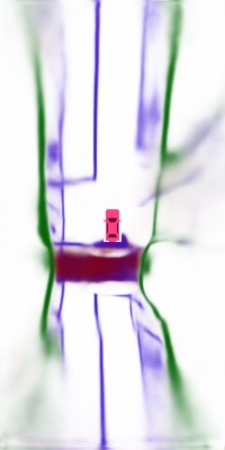}}
		\caption{pretrain + temp}\label{img:16_ssl_temp}
	\end{subfigure}

	\begin{subfigure}[t]{0.24\textwidth}
		\centering
		\rotatebox[origin=b]{270}{
			\includegraphics[width=0.5\linewidth]{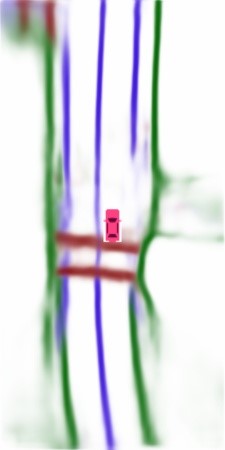}}
		\caption{ours(3ep pretrain + temp)}\label{img:16_ep3}
	\end{subfigure}
	\begin{subfigure}[t]{0.24\textwidth}
		\centering
		\rotatebox[origin=b]{270}{
			\includegraphics[width=0.5\linewidth]{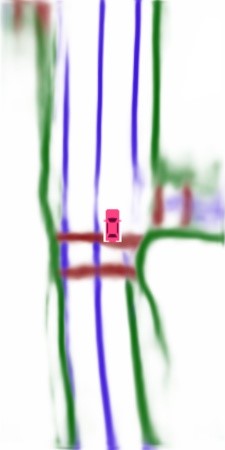}}
		\caption{ours(22ep pretrain + temp)}\label{img:16_ssl_temp_ep24}
	\end{subfigure}
	\begin{subfigure}[t]{0.24\textwidth}
		\centering
		\rotatebox[origin=b]{270}{
			\includegraphics[width=0.5\linewidth]{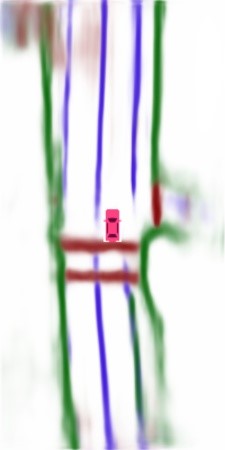}}
		\caption{supervised baseline} \label{img:16_baseline1image}
	\end{subfigure}
	\begin{subfigure}[t]{0.24\textwidth}
		\centering
		\rotatebox[origin=b]{270}{
			\includegraphics[width=0.5\linewidth]{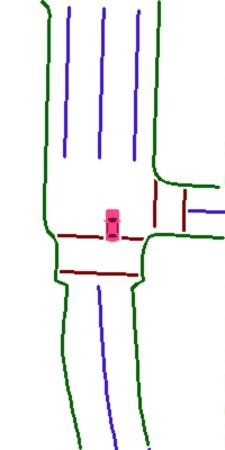}}
		\caption{nuScenes ground truth} \label{img:gt}
	\end{subfigure}
	\begin{subfigure}[t]{0.48\textwidth}
		\centering
		\includegraphics[width=\linewidth]{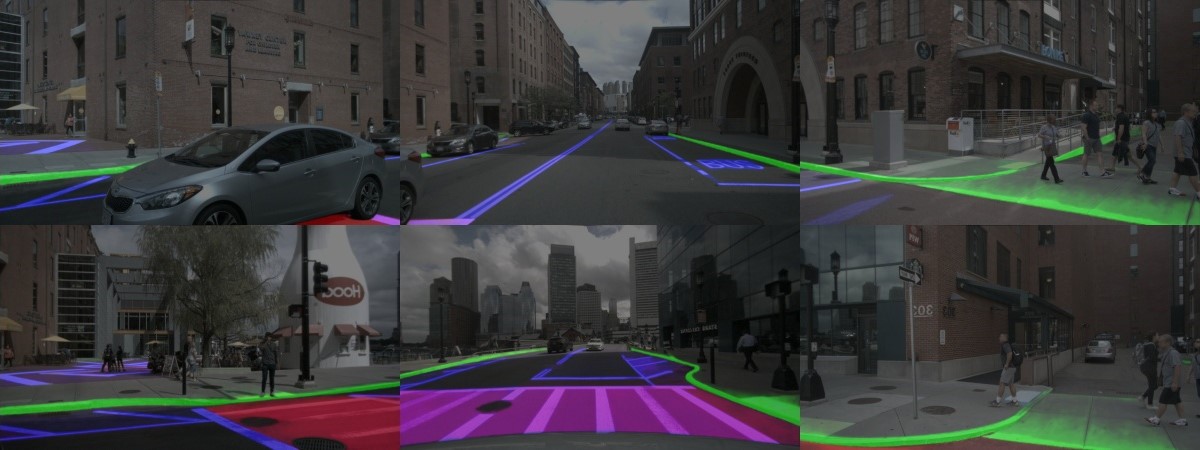}
		\caption{Camera images overlaid by pseudo ground truth.} \label{16_inout}
	\end{subfigure}
	\caption{Results of our self-supervised pretraining (pretrain), the temporal loss mechanism (temp) and our two phase strategy including temporal loss pretraining and supervised fine-tuning (ours($x$ epochs pretraining)). \textcolor{green}{road boundary}; \textcolor{blue}{lanes}; \textcolor{red}{crosswalk}.}\label{sample16}
\end{figure}
\subsection{Pretraining results}\label{quantitative_results}
In the first part of \cref{pretraining_length} we compare the pretraining with and without the temporal loss mechanism against the supervised baseline.
Additionally, we visualize example results of both pretrainings in (\cref{img:16_ssl,img:16_ssl_temp}).
The results indicate that the temporal loss mechanism provides slight improvements by $+0.7pp$ \ac{miou}60 over the pretraining without it.
Between the two pretrainings we observe the largest gain in the crosswalk class ($+2.2$ pp \ac{iou}), attributable to blind-spot crosswalk-artifacts near the ego vehicle that are partially mitigated by the temporal regularization \cref{img:16_ssl,img:16_ssl_temp}. These artifacts arise from misclassification in regions not visible in the current camera views but intermittently exposed through ego-motion. The temporal loss leverages this exposure and can suppress misclassified predictions. Consistent with this explanation, the relative improvements in \ac{miou}20 and \ac{miou}40 ($+1.1$pp and $+1.0$pp) are higher than at $60$m, since the affected blind-spot area occupies a larger proportion of the near- and mid-range evaluation regions. Despite this, the overall appearance of the predicted \ac{bev} maps with temporal loss appears less certain as it smooths detailed structures especially in further distances. This observation may lead to the fact that each sample has two ground truth maps (current and previous time step) to fulfill the temporal loss, which could introduce ambiguity during training.
Both pretraining variants remain below the supervised baseline ($-15.6$pp and $-14.9$pp \ac{miou}60). This gap is expected for two reasons: First, pseudo labels in the camera view may be affected by occlusions and incomplete visibility. Second, the pseudo-labeling guidelines differ from nuScenes map annotation policies (\cref{extension}). As illustrated in \cref{img:16_ssl,img:16_ssl_temp}, the pretraining outputs diverge from nuScenes ground truth  (\cref{img:gt}) yet match the pseudo labels (\cref{16_inout}). They include fine details (restricted zones, stop lines, bus symbols) absent from nuScenes annotations, reducing the measured \ac{miou} despite qualitatively plausible predictions. Although not generalizing well to the nuScenes ground truth, these results offering good priors for road marking structures that can be leveraged in subsequent fine-tuning as evaluated in the next section.

\subsection{Two phase training results}\label{qualitative_results}
\begin{figure}[htbp]
	\centering
	\includegraphics[width=0.98\linewidth]{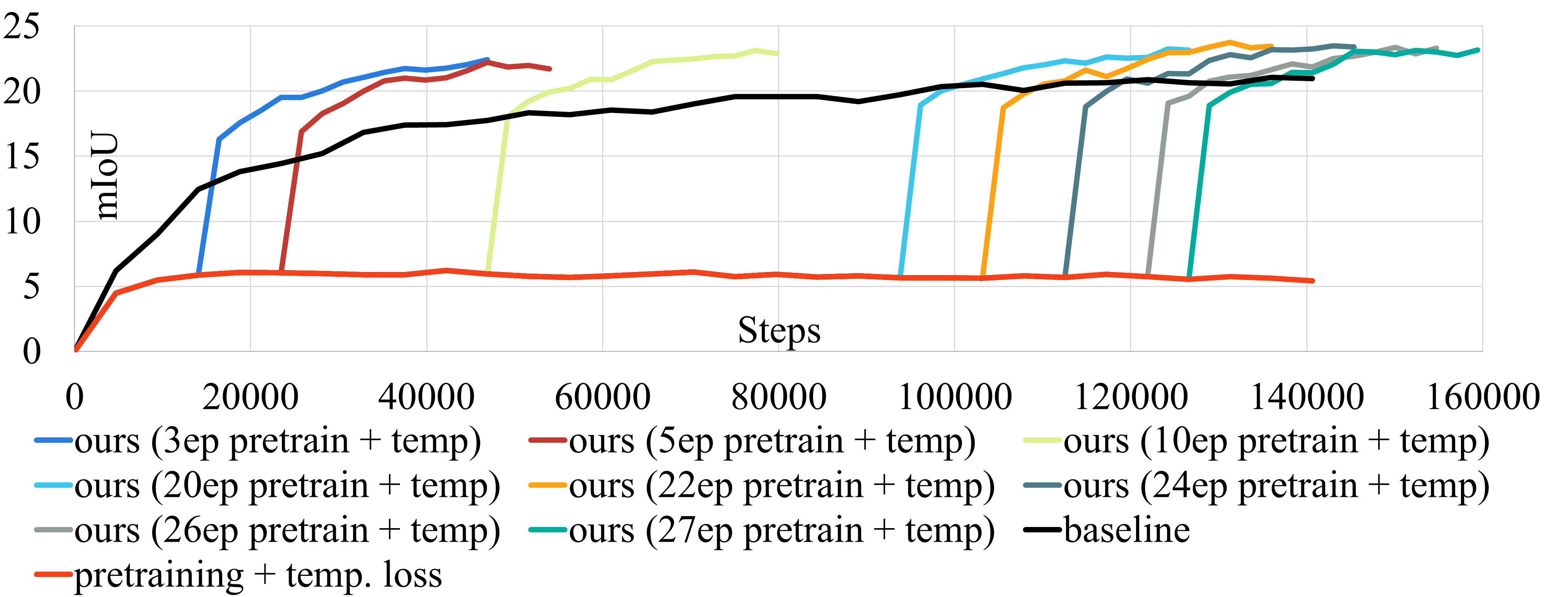}
	\caption{\ac{miou} progress evaluated on nuScenes ground truth for the supervised baseline, self-supervised pretraining, and our two phase training with different pretraining lengths including the temporal loss.}
	\label{fintune_miou}
\end{figure}

In the main part of \cref{pretraining_length} we present the detailed results of our two phase trainings.
In contrast to our pretrainings, each of our two phase trainings surpass the baseline in almost every metric. This demonstrates the effectiveness of our self-supervised pretraining combined with supervised fine-tuning as it can leverage the advantages from both perspectives while mitigating their individual limitations. For example, rich foreground details in camera perspective as presented by \cref{pixelrelation} and non occluded annotations during fine-tuning. The results indicate that a longer pretraining phase generally leads to better performance in the fine-tuning phase. The best overall result is achieved with a pretraining length of $22$ epochs, resulting in an \ac{miou}$60$ of $23.5pp$. Beyond this point longer fine-tuning may be required as to balance the two training phases.
The best predicted class varies with pretraining length: the road boundary class peaks with $27$ epochs, lane markings with $22$ epochs, and crosswalks with $26$ epochs of pretraining.
This suggests that different classes may benefit from different pretraining lengths, potentially due to their varying complexity and occlusion frequency in the pseudo ground truth. Road boundaries are often occluded by parked or moving vehicles, reducing effective supervision and may require longer pretraining. Crosswalks, though less frequent, may benefit from the different pseudo-label representation as filled areas, achieving the largest gain ($+4.5$ pp \ac{iou}).
However, it is worth noting that even with a short pretraining length of $3$ epochs, the model still outperforms the supervised baseline by $+1.4pp$ in \ac{miou}$60$ requiring only $1/3$ of the baseline training length. This suggests that the self-supervised pretraining is effective even with limited training time.\\
\textbf{Temporal loss.} We also report two-phase results without the temporal loss for the shortest (ours(3ep pretrain)) and best (ours(22ep pretrain)) pretraining settings (\cref{pretraining_length}). The small differences after fine-tuning indicate that the temporal loss is less effective on final performance as it is primarily intended to aid occluded road markings and blind spots during pretraining which appears less critical once the fine-tuning is applied.\\
\textbf{mIoU Progress.} To give more insights into the training dynamics of our two phase training, we visualize the \ac{miou}$60$ progress over the complete training length in \cref{fintune_miou}. It shows, the \acp{miou}$60$ of the baseline (black graph) the self-supervised pretraining (red graph) and the different compositions of pretraining and fine-tuning. The almost vertical steps are effects of the phase change between pretraining and fine-tuning and marking the beginning of fine-tuning. The steep increase in \ac{miou} within the first couple of fine-tuning steps, is observed for all fine-tuning phases helping to pass the baseline with in a few epochs of fine-tuning, regardless of the pretraining length. This underlines the effectiveness of the pretraining. Although, these steps appear smaller for short pretraining lengths where the pretraining has not yet converged (ours(3ep pretrain + temp) blue). Despite that, training with the longest pretraining phase (ours(27ep pretrain + temp) green) crosses the baseline just before the baseline training stops.\\
\textbf{Qualitative results.} The qualitative results support the aforementioned improvements achieved through our two phase training approach.
Although the pure pretraining (\cref{img:16_ssl,img:16_ssl_temp}) show some discrepancies compared to the nuScenes ground truth (\cref{img:gt}) it predicts detailed road marking structures. These predictions are significantly improved by our shorter two-phase training approach with $3$ epochs of pretraining (\cref{img:16_ep3}), approaching the appearance of the ground truth map. Lastly, with $22$ epochs of pretraining (\cref{img:16_ssl_temp_ep24}), the model produces higher quality \ac{bev} maps that closely resemble the ground truth (bottom left) even in further distances as well as details like the bottom intake. This underlines the quantitative findings for the longer pretraining lengths.

\section{Discussion and Limitation}\label{discussion}
Our experiments have shown that our self-supervised pretraining alone underperforms the fully supervised baseline, yet it provides a transferable camera-to-\ac{bev} feature mapping that significantly accelerates and amplifies supervised fine-tuning while reducing required \ac{bev} annotations.The increase within the first fine-tuning steps suggests that geometric projection and feature lifting are largely learned during pretraining, allowing supervised fine-tuning to focus on label alignment rather than spatial reasoning.\\
However, we have observed two main limitations in our approach: Firstly, the temporal consistency loss reduce artifacts but can over-smooth predictions which might be caused by ambiguity. We will address this by refined temporal objectives, add sparse map prediction, and incorporate explicit occlusion detection in future work. Secondly, a mismatch persists between the pseudo ground truth and the nuScenes \ac{bev} evaluation maps, due to representation gaps, occlusions, and annotation inconsistencies as depicted in \cref{lanethickness} (red crosswalks and orange marks). We plan to optimize pseudo ground truth generation to better match evaluation labels and investigate further reductions in \ac{bev} supervision in future studies. We also plan to extend the framework to dynamic object detection to enrich the \ac{bev} outputs.

\section{Conclusion}\label{conclusion}
To summarize, we have presented a novel two phase training approach for \ac{bev} semantic segmentation of fine-grained road markings that effectively leverages self-supervised pretraining on camera perspective segmentation images followed by a reduced supervised fine-tuning on $50\%$ \ac{bev} ground truth. Our experiments on nuScenes show significant improvement in three key aspects: it enhances \ac{bev} prediction performance by $+2.5pp$ in \ac{miou}, halves \ac{bev} label requirements, and can decrease the overall training time to $1/3$ while still yielding a $+1.4pp$ in \ac{miou}.
Future work will focus on improving pseudo ground truth generation, and extend to object detection tasks to further enrich \ac{bev} predictions.

\bibliographystyle{IEEEtran}
\bibliography{references01}
\end{document}